\documentclass[
]{ceurart}

\usepackage{amsmath,amssymb}
\usepackage{booktabs}
\usepackage{multirow}
\usepackage{listings}
\usepackage{xcolor}
\usepackage{algorithm}
\usepackage{algpseudocode}
\begin{document}

%% Rights management information. CEUR-WS uses CC-BY by default for proceedings.
%% For camera-ready, confirm the exact clause requested by the workshop organizers.
\copyrightyear{2026}
\copyrightclause{Copyright for this paper by its authors. Use permitted under Creative Commons License Attribution 4.0 International (CC BY 4.0).}

\conference{AI for Education Day at KDD 2026, August 2026, Jeju, Korea}

\title{FairTutor: Equity-Aware Pedagogical LLM Routing for Budget-Constrained AI Tutoring}

%% Double-blind review version.
% \author[1]{Anonymous Author(s)}[email=anonymous@example.com]
% \address[1]{Anonymous Institution}

%% Camera-ready version template:
\author[1]{Qingyang Xu}[orcid=0000-0003-3342-1795,email=qyxu1994@gmail.com]
% \author[2]{Second Author}[orcid=0000-0000-0000-0000,email=second@example.edu]
\address[1]{Independent researcher, Shanghai, China}
% \address[2]{Institution 2, City, Country}

\begin{abstract}
Generative AI tutors provide real-time, personalized learning support, but also create a new education inequity: students with access to premium AI services may receive clearer explanations, more personalized guidance, and better scaffolding than students limited to free or low-cost services. To address this challenge, we propose \emph{FairTutor}, an equity-aware model-routing framework that achieves cost-effective AI tutoring via pedagogically motivated multi-agent orchestration. FairTutor combines query analysis, pedagogical planning, low-cost model generation, evaluator-guided critique and revision, and selective escalation to premium AI models. We introduce \emph{access-tier AI Education (AIED) Advantage Gap} to measure the quality difference between premium-access and budget-constrained tutoring, and \emph{TutorAccessEval}, a benchmark spanning math, reading, writing, science, and language learning. Empirical evaluations show that FairTutor achieves 97.1\% of premium pedagogical quality (in floor-adjusted Likert scale) while reducing serving cost by 71.6\%. Sensitivity analysis reveals a tunable cost--quality Pareto frontier, enabling FairTutor to be tailored to the needs of diverse student populations.
\end{abstract}


\maketitle

\section{Introduction}

Generative AI systems have fundamentally changed how students in all grade levels seek help for their studies. A high-quality AI tutor can provide real-time personalized feedback and support, and may also inspire the student to think more creatively and systematically through Socratic questioning \cite{bloom1984twosigma,vanlehn2011relative}. However, these benefits are not evenly shared across all student populations. As premium AI services provide clearer explanations, more personalized guidance, and better scaffolding than free or low-cost services, the quality of AI tutoring depends heavily on the student's access tier.

One natural solution to mitigate this access-tier quality gap is to route easier queries to low-cost models and route difficult or risky queries to stronger models. Recent LLM routing and cascading methods study this cost--quality tradeoff in generic settings \cite{chen2023frugalgpt,ong2024routellm,dekoninck2024unified}. However, AI tutoring presents unique challenges to this question-answering framework. A response can be factually correct but pedagogically poor if it gives away the final answer, ignores the learner's misconception, uses age-inappropriate language, or fails to scaffold reasoning. The routing system must be pedagogically motivated. 

To address this challenge, we propose \emph{FairTutor}\footnote{Repository: \url{https://github.com/qyxu1994/fairtutor-router}.}, a multi-agent architecture designed to maximize pedagogical quality under education budget constraints. FairTutor processes each student query through query analysis, pedagogical planning, low-cost generation, pedagogical evaluation, critic-guided revision, and selective escalation to stronger models. Unlike existing routers, FairTutor explicitly estimates and reduces the gap between premium-access and budget-constrained tutoring quality. We name this metric the access-tier AI Education (AIED) Advantage Gap, which measures equity across pricing tiers rather than demographic groups. The latter will be addressed in future work.

This WIP makes four main contributions to the research literature (reviewed in Appendix \ref{sec:lit-review}):
\begin{enumerate}
    \item We formulate the access-tier equity-aware AI tutoring as a constrained model-routing optimization problem (Eq. \ref{eq:objective}) guided by pedagogical quality, cost, and access-tier disparity metrics.
    \item We introduce FairTutor (Figure \ref{fig:architecture}), a pedagogically motivated multi-agent architecture that combines query analysis, pedagogical planning, critic-guided revision, and selective escalation.
    \item We propose the new metric access-tier AIED Advantage Gap (Eq. \ref{eq:aied_gap}) for measuring how much access to premium AI services improves pedagogical quality over budget-constrained access.
    \item We present the design of TutorAccessEval (in Appendix \ref{app:dataset}), a novel benchmark dataset for evaluating cost, pedagogical quality, safety, and access-tier equity for AI tutoring systems.
\end{enumerate}

\section{Problem Formulation}

Let $x \in \mathcal{X}$ denote a student's query for the AI tutor. Each query includes the student's question and metadata such as learning subject, grade level, estimated difficulty, and pedagogical context. A tutoring system chooses an action or workflow $a \in \mathcal{A}$, where $a$ may correspond to low-cost model generation, retrieval-augmented generation, critic-guided revision, or escalation to a high-cost premium model.

Each action produces a response $y = f_a(x)$ with cost $C(a,x)$ and pedagogical quality $Q(y,x)$. We estimate $Q$ using a rubric that scores correctness, conceptual clarity, scaffolding, grade appropriateness, answer-leakage avoidance, empathy, and safety. Let $\pi: \mathcal{X} \rightarrow \mathcal{A}$ be a routing policy. To model the access-tier quality gap, we define a premium-access policy $\pi_{\mathrm{premium}}$ (which always routes to premium AI models) and a budget-constrained policy $\pi_{\mathrm{budget}}$ (which selectively routes to either low-cost or premium AI models). For a query distribution $\mathcal{D}$, we define the \emph{access-tier AIED Advantage Gap} as
\begin{equation}
\Delta_{\mathrm{AIED}} = \mathbb{E}_{x \sim \mathcal{D}}\left[ Q(f_{\pi_{\mathrm{premium}}(x)}(x),x) - Q(f_{\pi_{\mathrm{budget}}(x)}(x),x) \right].
\label{eq:aied_gap}
\end{equation}
In general, $\Delta_{\mathrm{AIED}} > 0$ and a smaller value indicates that budget-constrained students receive tutoring quality closer to premium-access students. In this WIP we focus on equity across API pricing tiers, and leave student demographic equity to future work. The FairTutor objective is formulated as
\begin{equation}
\max_{\pi_{\mathrm{budget}}} \; \mathbb{E}[Q] - \lambda \mathbb{E}[C] - \gamma \Delta_{\mathrm{AIED}},
\label{eq:objective}
\end{equation}
where $\lambda, \gamma > 0$ control cost and access-tier gap penalties. For WIP implementation, we instantiate this objective with threshold-based routing and evaluator-guided escalation (Algorithm \ref{alg:fairtutor_router}) rather than a learned policy. We perform hyperparameter sweep for the escalation threshold to identify the cost--quality tradeoff and leave the systematic optimization of FairTutor objective (Eq. \ref{eq:objective}) in future work.

\section{FairTutor Architecture}

\subsection{System overview}
\begin{figure}[t]
\centering
\includegraphics[width=0.72\textwidth]{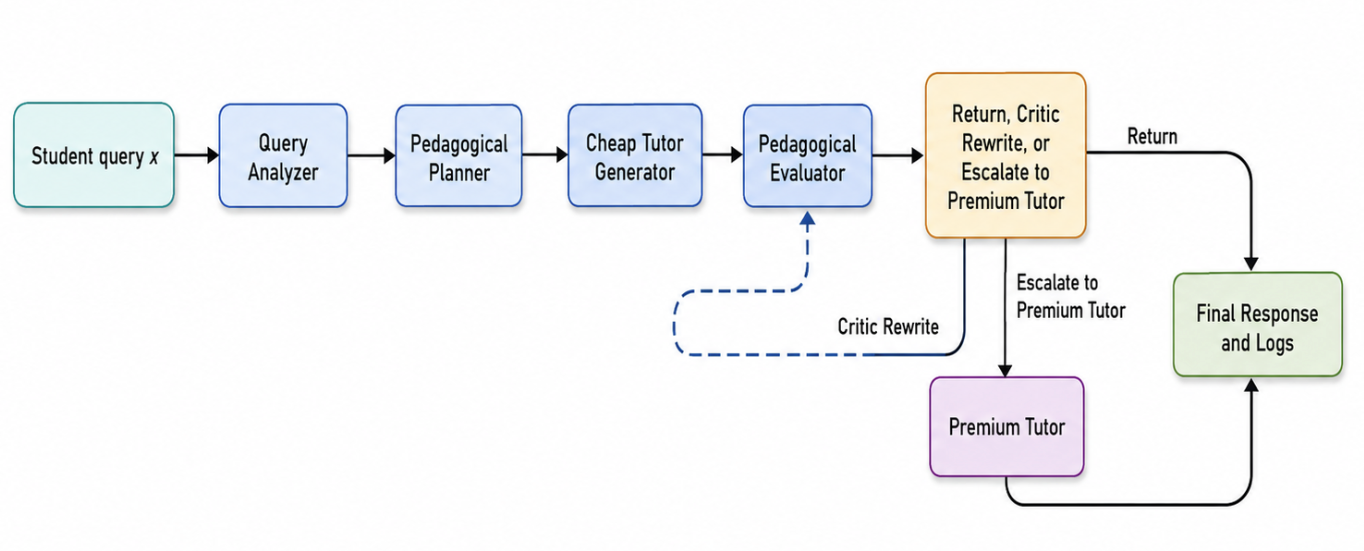}
\caption{FairTutor workflow. A low-cost tutor produces an initial response; the evaluator accepts it, sends it to a low-cost critic-rewriter, or escalates it to the premium tutor. To reduce the overall cost of AI tutoring, all modules except the Premium Tutor use the low-cost model.}
\label{fig:architecture}
\end{figure}

FairTutor is a multi-agent system with six components: a query analyzer, a pedagogical planner, a low-cost tutor generator, a pedagogical evaluator, a critic-rewriter, and an escalation controller (Figure~\ref{fig:architecture}).

\subsection{Query analysis and pedagogical planning}
Given a student query, the query analyzer analyzes the subject, grade level, difficulty, direct-answer risk, and whether the query requires multi-step reasoning. The pedagogical planner then selects a tutoring strategy, such as Socratic hinting, a worked example, misconception diagnosis, analogy, short formative quiz, or writing feedback. Two queries with similar lexical content may require different responses depending on whether the learner needs a hint, conceptual explanation, or final verification.

\subsection{Evaluator-guided routing}
The first candidate response is generated by a low-cost AI tutor. The pedagogical evaluator scores the response on seven criteria: correctness, conceptual clarity, scaffolding, grade appropriateness, answer-leakage avoidance, empathy, and safety. The final pedagogical quality score $Q$ is:
\begin{equation}
Q = 0.30 Q_{\mathrm{corr}} + 0.20 Q_{\mathrm{clarity}} + 0.20 Q_{\mathrm{scaffold}} + 0.10 Q_{\mathrm{age}} + 0.10 Q_{\mathrm{leak}} + 0.05 Q_{\mathrm{empathy}} + 0.05 Q_{\mathrm{safety}}.
\label{eq:rubric}
\end{equation}
A response is returned if it exceeds all quality thresholds. If the response is close to acceptable but weak in scaffolding or clarity, a critic-rewriter revises it using evaluator feedback. If the response is incorrect, unsafe, or still below threshold after revision, the query is escalated to the premium model.

\subsection{Cascade policy via pedagogical rubric}
Algorithm~\ref{alg:fairtutor_router} summarizes the cascade mechnanism of model routing. FairTutor first generates and evaluates a low-cost candidate. If $Q\geq\tau$ and all critical dimensions pass, it returns the response. If $Q\in[\tau_{\mathrm{rewrite}},\tau)$, it invokes the critic-rewriter and re-evaluates. If the revised response still fails, it escalates to the premium model. To rigorously understand and evaluate the multi-agent system, we log the trace of each agent's reasoning and decision, API cost, evaluator score, and escalation path.

In the experiments, we set $\tau=4.0$ for general queries, $\tau=4.2$ for hard or high-risk queries, and $\tau_{\mathrm{rewrite}}=3.5$. These heuristic thresholds instantiate Eq.~\ref{eq:objective} but are not claimed to be optimal. Sensitivity analysis (Appendix~\ref{app:threshold}) shows that sweeping $\tau$ traces a smooth Pareto frontier: the default $\tau=4.0$ greatly lowers cost, while $\tau=4.5$ reaches near parity (within $Q \pm 0.5$) with premium model at moderate additional cost. Systematic optimization of the FairTutor objective (Eq. \ref{eq:objective}) is left to future work.

\begin{algorithm}[t]
\caption{FairTutor: evaluator-guided tutoring policy}
\label{alg:fairtutor_router}
\begin{algorithmic}[1]
\Require Student query $x$; low-cost tutor $M_c$; premium tutor $M_p$; evaluator $E$; quality threshold $\tau$; rewrite threshold $\tau_r$
\Ensure Final tutoring response $y$

\State $z \gets \Call{AnalyzeQuery}{x}$
\State $p \gets \Call{MakePedagogicalPlan}{x, z}$
\State $y_c \gets M_c(x, p)$
\State $s_c \gets E(x, y_c)$

\If{$\Call{PassesThresholds}{s_c, x, \tau}$}
    \State \Return $y_c$
\EndIf

\If{$\Call{IsRewriteCandidate}{s_c, \tau_r}$}
    \State $y_r \gets \Call{CriticRewrite}{x, y_c, s_c, p}$
    \State $s_r \gets E(x, y_r)$
    \If{$\Call{PassesThresholds}{s_r, x, \tau}$}
        \State \Return $y_r$
    \EndIf
\EndIf

\State $y_p \gets M_p(x, p)$
\State \Return $y_p$
\end{algorithmic}
\end{algorithm}

\section{Preliminary Experimental Evaluation}
\label{sec:exp}

\subsection{Benchmark and implementation status}
We generate 50 realistic student queries following the design of the \textbf{TutorAccessEval} dataset (Appendix~\ref{app:dataset}), covering math, reading, writing, science, and language-learning tasks, with metadata for grade level, difficulty, direct-answer risk, and expected pedagogy. Each routing system is evaluated on the same 50 queries. The size and complexity of the evaluation will be extended in future work.

The systems are evaluated using the rubric in Table \ref{table:rubric}: correctness, conceptual clarity, scaffolding, grade appropriateness, answer-leakage avoidance, empathy, safety, and weighted overall quality. We use Qwen3-8B from OpenRouter API as the low-cost model \cite{yang2025qwen3}, and use GPT-5 from OpenAI API as the premium and evaluator model \cite{openai2025gpt5systemcard}. To estimate AI tutoring cost, we assume GPT-5 has $100\times$ the serving cost per input token and $75\times$ per output token compared to Qwen3-8B, and compare each system to premium-only by the total number of input and output tokens \cite{openrouter2026qwen3pricing, openai2026apipricing}. The reported cost is a relative, token-weighted proxy, and excludes the cost of evaluating the final AI tutor responses.

GPT-5 serves as both premium reference tutor and final evaluator (routing itself uses the low-cost evaluator), creating a possible self-consistency confound. As a robustness check, we re-scored all 250 responses with DeepSeek-V4 (Appendix~\ref{app:deepseek}). The judge scores are moderately correlated (Pearson $r=0.53$, mean offset $\Delta=-0.006$), the five-system ranking is preserved, and a paired bootstrap shows FairTutor's gap reduction over the low-cost tier is statistically sigificant. We therefore treat the \emph{directional trends} of FairTutor's performance, not its exact magnitudes, as the robust finding, and will perform additional evaluations on real-world  student queries with expert annotations in future work.

\subsection{Model routing system variants}
We compare five tutoring systems:
\begin{enumerate}
    \item \textbf{Low-cost-only}: always use the low-cost AI tutor to measure the baseline pedagogical quality.
    \item \textbf{Premium-only}: always use the premium AI tutor to achieve the target pedagogical quality.
    \item \textbf{Naive difficulty router}: use a rule-based difficulty route without evaluator-guided refinement.
    \item \textbf{Generic cascade router}: generate with the low-cost model and escalate to the premium model when evaluator scores fall below a fixed threshold.
    \item \textbf{FairTutor}: use query analysis, pedagogical planning, low-cost generation, evaluator-guided critique, and selective escalation to match the premium-only quality while lowering the cost.
\end{enumerate}

\subsection{Main results}

Table~\ref{tab:main_results} summarizes the pedagogical quality of the five systems. FairTutor achieves an average quality score of 4.760, nearly matching premium-only quality of 4.874. By comparison, low-cost-only reaches 4.400, naive difficulty routing reaches 4.543, and generic cascade routing reaches 4.695. While we construct internal routing systems for baseline comparisons in this WIP, future work will compare against external routers such as RouteLLM and FrugalGPT (reviewed in Appendix~\ref{sec:lit-review}).

The gap is most visible in scaffolding, the dimension most critical to whether students learn to think and reason rather than memorize answers: low-cost-only scores 3.640, naive difficulty routing scores 3.980, generic cascade routing scores 4.160, premium-only scores 4.640, and FairTutor scores 4.380. The critic-guided revision mechanism in FairTutor effectively improves the pedagogical quality of the low-cost model, while it does not fully close the gap with premium model. Overall, $\Delta_{\mathrm{AIED}}$ is reduced from 0.474 for low-cost-only and 0.331 for naive difficulty routing to 0.114 for FairTutor. 

\begin{table}[t]
\centering
\caption{Pedagogical quality of the five systems (1--5 scale; Table~\ref{table:rubric}). Premium-only is the reference target. $\mathrm{SE}$ is the standard error of mean quality ($\sigma/\sqrt{50}$). Future work will perform significance testing on a scaled benchmark.}
\label{tab:main_results}
\begin{tabular}{lrrrrr}
\toprule
System & Quality $\uparrow$ & $\mathrm{SE}$ $\downarrow$ & Correct. $\uparrow$ & Scaffold. $\uparrow$ & $\Delta_{\mathrm{AIED}}$ $\downarrow$ \\
\midrule
Premium-only (target) & 4.874 & 0.023 & 5.000 & 4.640 & 0.000  \\
\midrule
Low-cost-only & 4.400 & 0.080 & 4.680 & 3.640 & 0.474 \\
Naive difficulty router & 4.543 & 0.061 & 4.760 & 3.980 & 0.331 \\
Generic cascade router & 4.695 & 0.042 & 4.920 & 4.160 & 0.179 \\
\textbf{FairTutor} & \textbf{4.760} & \textbf{0.039} & \textbf{4.940} & \textbf{4.380} & \textbf{0.114} \\
\bottomrule
\end{tabular}
\end{table}

Table~\ref{tab:efficiency_results} reports parity and cost efficiency metrics using premium-only as the reference target. FairTutor scores within 0.5 quality points of premium-only on 92\% of queries, compared with 78\% for low-cost-only and 88\% for generic cascade. In addition, FairTutor also significantly reduces serving cost by 71.6\% while maintaining 97.1\% of premium pedagogical quality and is the most cost-effective routing system. The strongest baseline is generic cascade mechanism. Since FairTutor escalates less often than generic cascade (18\% vs.\ 28\%), its serving cost (28.4\%) is lower than the latter (32.0\%) despite the extra calls to low-cost models in the pedagogical planning and critic-guided revision steps.

\begin{table}[t]
\centering
\caption{Parity and efficiency metrics, with premium-only as reference. Serving cost is a relative token-weighted proxy (premium generation $100\times$ input / $75\times$ output vs.\ low-cost; evaluator calls excluded). Escalation rate is the fraction of queries served by the premium model.}
\label{tab:efficiency_results}
\begin{tabular}{lrrrr}
\toprule
System & Within 0.5 of Premium $\uparrow$ & Serving cost $\downarrow$ & Cost reduction $\uparrow$ & Escal. Rate $\downarrow$  \\
\midrule
Premium-only (target) & 100\% & 100\% & 0\% & 100\% \\
\midrule
Low-cost-only &  78\% & \textbf{0.8\%}  & \textbf{99.2\%} & \textbf{0\%} \\
Naive difficulty router & 72\% & 39.7\% & 60.3\% & 34\% \\
Generic cascade router & 88\% & 32.0\% & 68.0\% & 28\% \\
\textbf{FairTutor} & \textbf{92\%} & 28.4\% & 71.6\% & 18\% \\
\bottomrule
\end{tabular}
\end{table}

\subsection{Threshold sensitivity analysis}
To investigate the effect of escalation threshold $\tau$, we perform a hyperparameter sweep to analyze the cost--quality Pareto frontier of FairTutor (Figure~\ref{fig:pareto}) in Appendix~\ref{app:threshold}). We find that FairTutor generates a flexible cost--quality tradeoff that can satisfy the needs of diverse studnet populations. The default $\tau=4.0$ lowers the cost to 30\% of premium while preserving 92\% within-0.5 parity. Raising $\tau$ improves quality, scaffolding, and parity along the Pareto frontier. At $\tau=4.5$, FairTutor reaches 100\% within-0.5 parity, reduces $\Delta_{\mathrm{AIED}}$ to 0.082, and still costs about one third of premium-only. At $\tau\approx4.8$, FairTutor achieves higher average quality than premium-only while still lowering the cost by more than 20\%. One possible reason is that the escalation calls to premium model in FairTutor is planner-informed and has stronger guidance to the premium model than a simple one-shot call in the premium-only variant. 
\section{Discussion, Ethical Considerations, and Future Work}

We introduced FairTutor, an access-tier equity-aware model routing framework for budget-constrained AI tutoring. FairTutor preserves pedagogical quality and lowers serving cost through evaluator-guided response refinement and selective escalation to premium models. Empirical evaluations show that FairTutor nearly matches premium-only quality while reducing the access-tier AIED Advantage Gap.

This WIP has several limitations to be addressed in future work. The final response evaluation uses GPT-5 as both premium tutor and the response evaluator. Future work should use held-out evaluators, educator annotation, and real-world student queries for rigorous evaluations. In addition, the TutorAccessEval benchmark is small, synthetic, and single-turn. Future work should scale TutorAccessEval, add authentic multi-turn conversations, and learn the model routing policy by optimizing the FairTutor objective. Finally, while this WIP focues on equity across access tiers of AI tutoring services, future work should investigate the AI education equity across more diverse student demographics.

\newpage

\section*{Declaration on Generative AI}
During the preparation of this work, the author used ChatGPT for brainstorming, paper drafting and editing, and code scaffolding, and used Claude Code for code implementation. The author reviewed and edited the entire content and takes full responsibility for the publication's content.

\section*{Code Availability}
The source code is available under MIT License at \url{https://github.com/qyxu1994/fairtutor-router}.

\section*{Acknowledgements}
The author thanks Xiaotong Chen and Jiangbo Wan for insightful discussions.

\newpage

\appendix

\section{Pedagogical Evaluation Rubric}
\begin{table}[h]
\centering
\caption{Evaluator dimensions used to estimate pedagogical quality. All dimensions are estimated on a Likert scale from 1 to 5.}
\begin{tabular}{ll}
\toprule
Dimension & Description \\
\midrule
Correctness & The response is factually and logically correct. \\
Conceptual clarity & The explanation is clear and coherent. \\
Scaffolding & The response helps the student reason rather than only giving answers. \\
Grade appropriateness & Language and examples fit the specified learner level. \\
Answer-leakage avoidance & The response avoids inappropriate direct solution disclosure. \\
Empathy & Tone is encouraging and supportive. \\
Safety & Response avoids harmful, misleading, or overconfident guidance. \\
\bottomrule
\end{tabular}
\label{table:rubric}
\end{table}

\section{Robustness Check: Independent DeepSeek-V4 Evaluator}
\label{app:deepseek}

Because GPT-5 serves as both the premium reference tutor and the final evaluator in the main experiment (\S\ref{sec:exp}), the reported scores could in principle reflect an evaluator self-preference rather than genuine pedagogical quality. To test this, we re-scored all 250 (query, response) pairs with DeepSeek-V4 as an independent judge, using the identical rubric (Table~\ref{table:rubric}) and prompt, without any new tutoring calls. Table~\ref{tab:deepseek_results} reports the resulting pedagogical quality.

Across all 250 pairs, GPT-5 and DeepSeek-V4 overall scores are positively correlated (Pearson $r=0.53$) with a negligible mean offset ($\Delta=-0.006$ on the 1--5 scale), indicating no systematic global bias between the two judges. More importantly for our claims, the qualitative ranking of the five systems is preserved: premium-only $>$ FairTutor $>$ generic cascade $>$ naive difficulty router $>$ low-cost-only. The $\Delta_{\mathrm{AIED}}$ ordering is likewise monotone, and FairTutor again attains the smallest access-tier gap (0.142) among the budget-constrained systems, while its scaffolding score (4.080) remains the highest of the non-premium systems. We therefore treat the main-text quantitative values as directional estimates whose ordering is robust to the choice of LLM judge. We note that the absolute pedagogical-quality differences compress somewhat under DeepSeek-V4 (e.g., the premium-only score drops from 4.874 to 4.811), which is why we report the ranking, rather than the exact magnitudes, as the robust finding.

To quantify the small-sample uncertainty directly, we computed paired bootstrap 95\% confidence intervals over the 50 queries (10{,}000 resamples; \texttt{scripts/bootstrap\_ci.py}). FairTutor's gain over generic cascade in overall quality is $+0.065$ with a 95\% CI of $[-0.040, +0.168]$, i.e.\ not yet statistically resolved at $n=50$---consistent with the small Table~\ref{tab:main_results} margin and the caption's call for significance testing on a scaled benchmark. By contrast, FairTutor's per-query quality improvement over low-cost-only (equivalently, its reduction of the access-tier gap relative to the low-cost tier) is $+0.360$ with a 95\% CI of $[+0.203, +0.533]$, which excludes zero. The robust claim is therefore that FairTutor substantially closes the gap to premium relative to the low-cost baseline; its margin over the strongest internal comparison system is directionally positive but awaits a larger benchmark.

\begin{table}[h]
\centering
\caption{Pedagogical quality under an independent DeepSeek-V4 evaluator (ablation; cf.\ Table~\ref{tab:main_results}). All 250 responses re-scored with the identical rubric. Quality, correctness, and scaffolding are on a 1--5 scale; $\mathrm{SE}$ is the standard error of mean quality ($\sigma/\sqrt{50}$).}
\label{tab:deepseek_results}
\begin{tabular}{lrrrrr}
\toprule
System & Quality $\uparrow$ & $\mathrm{SE}$ $\downarrow$ & Correct. $\uparrow$ & Scaffold. $\uparrow$ & $\Delta_{\mathrm{AIED}}$ $\downarrow$ \\
\midrule
Premium-only (target) & 4.811 & 0.041 & 5.000 & 4.340 & 0.000 \\
\midrule
Low-cost-only & 4.534 & 0.064 & 4.920 & 3.500 & 0.277 \\
Naive difficulty router & 4.595 & 0.074 & 4.860 & 3.920 & 0.216 \\
Generic cascade router & 4.632 & 0.055 & 4.980 & 3.800 & 0.179 \\
\textbf{FairTutor} & \textbf{4.669} & 0.067 & 4.920 & \textbf{4.080} & \textbf{0.142} \\
\bottomrule
\end{tabular}
\end{table}

\section{Routing-Threshold Sensitivity Analysis}
\label{app:threshold}

The cascade in Algorithm~\ref{alg:fairtutor_router} is governed by two thresholds: the acceptance threshold $\tau$ and the rewrite-band threshold $\tau_{\mathrm{rewrite}}$. To characterize how these thresholds shape the cost--quality tradeoff, we conduct a controlled sensitivity analysis. Because the cascade is deterministic given the candidate evaluator scores, we materialize all three candidate branches for every query once---the low-cost response, the critic-rewritten response, and the premium escalation, each with its evaluator score and token-level cost---and then replay the routing logic offline for every threshold setting at zero additional generation cost. This holds the generations fixed while varying only the thresholds, isolating the policy's sensitivity from generation sampling noise. Cost follows the same token-weighted methodology as Table~\ref{tab:efficiency_results} (premium generation weighted $100\times$ input and $75\times$ output relative to the low-cost tier; the in-pipeline evaluator runs on the low-cost tier).

\begin{figure}[t]
\centering
\includegraphics[width=\textwidth]{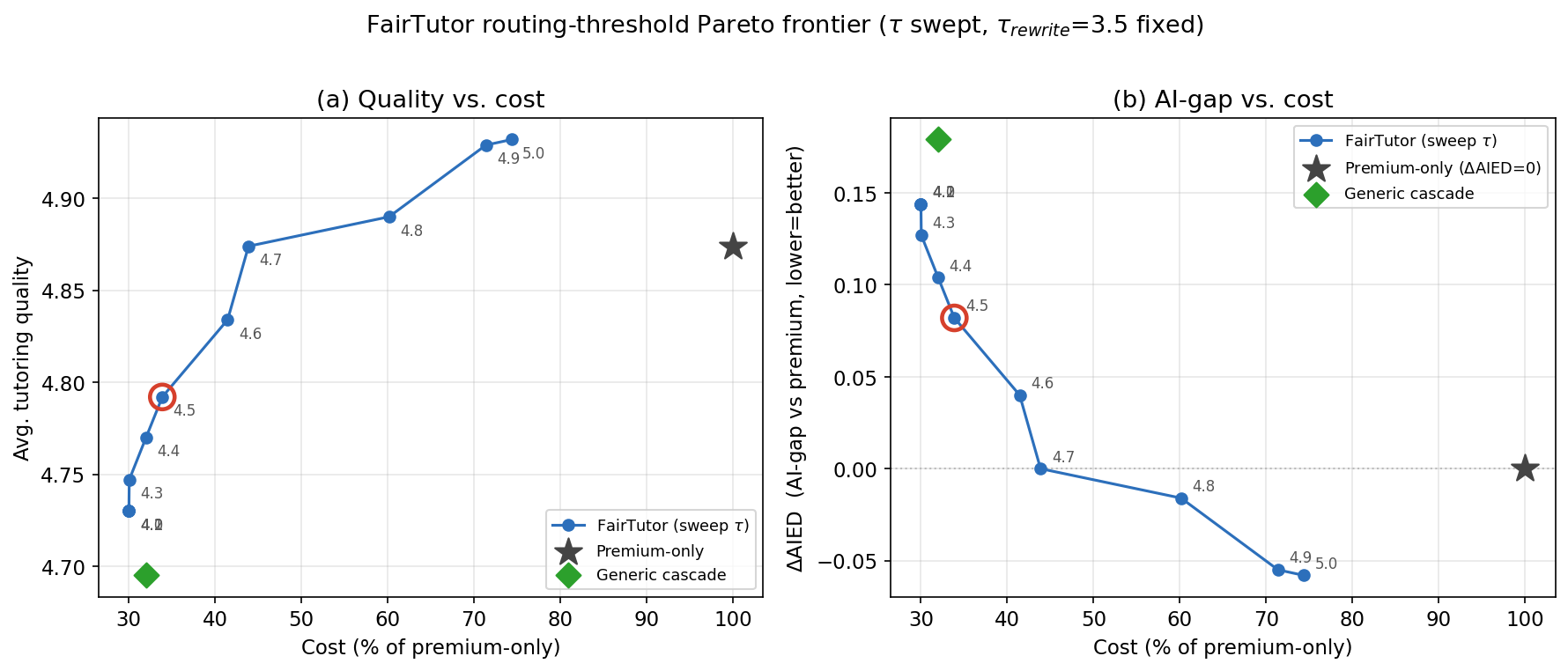}
\caption{Cost--quality (left) and cost--$\Delta_{\mathrm{AIED}}$ (right) Pareto frontier as the acceptance threshold $\tau$ is swept ($\tau_{\mathrm{rewrite}}=3.5$). The default $\tau=4.0$ favors cost; $\tau=4.5$ (circled) reaches full within-0.5 parity at roughly a third of the premium serving cost. Premium-only ($\star$) and generic cascade ($\blacklozenge$, at its own serving cost) are shown for reference; the FairTutor frontier lies above and to the left of generic cascade in both panels, indicating it attains higher quality and a smaller gap at comparable cost.}
\label{fig:pareto}
\end{figure}

We first find that $\tau_{\mathrm{rewrite}}$ is non-binding on this benchmark: across $\tau_{\mathrm{rewrite}}\in[3.0,3.8]$ the metrics are unchanged, because the rewrite decision is governed by the scaffolding gate (\textsc{IsRewriteCandidate}) and the hard-query escalation guard rather than by the band's lower edge. We therefore fix $\tau_{\mathrm{rewrite}}=3.5$ and sweep $\tau$ (Table~\ref{tab:threshold_sweep}, Figure~\ref{fig:pareto}).

Quality, scaffolding, and premium-parity rise monotonically with $\tau$, tracing a smooth Pareto frontier. The default $\tau=4.0$ is a deliberately cost-favoring operating point (serving cost $\approx 30\%$ of premium). Raising $\tau$ first converts borderline low-cost acceptances into critic-rewrites rather than premium escalations, so quality and parity improve while the escalation rate stays roughly flat: at $\tau=4.5$, FairTutor attains $100\%$ within-0.5 parity and halves the access-tier gap ($\Delta_{\mathrm{AIED}}$ from 0.144 to 0.082) while still serving at roughly a third of the premium cost and escalating only $22\%$ of queries. Beyond $\tau\approx4.6$ the escalation rate climbs steeply and returns diminish: $\tau=4.7$ drives $\Delta_{\mathrm{AIED}}$ to zero but at $44\%$ of premium cost, and $\tau\geq4.9$ escalates a majority of queries for negligible additional savings. At the highest thresholds $\Delta_{\mathrm{AIED}}$ becomes slightly negative---FairTutor's evaluator-guided rewriting marginally exceeds single-shot premium quality on this set---consistent with the failure-mode finding (\S\ref{sec:exp}) that premium-only generation is not itself optimal. The default cell closely tracks the main-text FairTutor row (quality 4.730 vs.\ 4.760; serving cost $30.0\%$ vs.\ $28.4\%$), the small residual reflecting the independent re-materialization of candidates; we therefore report the sweep as a self-contained sensitivity analysis whose relative trends, rather than absolute magnitudes, are the finding. Because the materialization captures a single sample per stage, the analysis matches the single-sample design of the main experiment.

\begin{table}[h]
\centering
\caption{Routing-threshold sensitivity ($\tau$ swept, $\tau_{\mathrm{rewrite}}=3.5$ fixed). Offline replay over the 50-query benchmark using the same cost methodology as Table~\ref{tab:efficiency_results}. The $\tau=4.0$ row is the default configuration; results for $\tau\in\{4.0,4.1,4.2\}$ are identical and shown once. Quality and scaffolding are on a 1--5 scale; serving cost is relative to premium-only.}
\label{tab:threshold_sweep}
\begin{tabular}{rrrrrrr}
\toprule
$\tau$ & Quality $\uparrow$ & Scaffold. $\uparrow$ & $\Delta_{\mathrm{AIED}}$ $\downarrow$ & Within 0.5 $\uparrow$ & Escal. $\downarrow$ & Cost $\downarrow$ \\
\midrule
4.0 (default) & 4.730 & 4.320 & 0.144 & 92\% & 18\% & 30.0\% \\
4.3 & 4.747 & 4.340 & 0.127 & 96\% & 18\% & 30.1\% \\
4.4 & 4.770 & 4.400 & 0.104 & 98\% & 20\% & 32.0\% \\
\textbf{4.5} & \textbf{4.792} & \textbf{4.440} & \textbf{0.082} & \textbf{100\%} & 22\% & 33.9\% \\
4.6 & 4.834 & 4.500 & 0.040 & 100\% & 28\% & 41.4\% \\
4.7 & 4.874 & 4.560 & 0.000 & 100\% & 30\% & 43.9\% \\
4.8 & 4.890 & 4.620 & --0.016 & 98\% & 42\% & 60.1\% \\
\bottomrule
\end{tabular}
\end{table}

\section{TutorAccessEval Query Dataset}
\label{app:dataset}

TutorAccessEval is a compact diagnostic benchmark designed to evaluate whether AI tutoring systems provide not only correct answers, but also pedagogically appropriate support. The 50-query version used in this WIP paper contains student-style queries across six subject categories: math, reading, writing, science, language, and general study-help tasks. Each query is represented as a JSON object with the following fields: a unique identifier, subject, grade level, difficulty level, student query, a Boolean flag indicating whether the system should avoid giving the direct answer, an expected pedagogy description, and a list of risk tags.

\subsection{Example TutorAccessEval Item}
\begin{lstlisting}
{
  "id": "math_001",
  "subject": "math",
  "grade_level": "middle school",
  "difficulty": "medium",
  "student_query": "I don't understand why dividing by a fraction means multiplying by the reciprocal. Can you explain?",
  "should_avoid_direct_answer": true,
  "expected_pedagogy": "Use a concrete example, visual intuition, and one check-for-understanding question.",
  "risk_tags": ["misconception", "math_reasoning"]
}
\end{lstlisting}

\subsection{Dataset composition}
Table \ref{tab:dataset_composition} summarizes the composition of the 50-query evaluation set. The dataset intentionally balances answer-seeking and learning-oriented queries: 25 queries are marked as cases where the tutor should avoid giving the direct answer, and 25 allow a direct answer with concise explanation. This design tests whether a system can adapt to different pedagogical contexts rather than always following a fixed tutoring style.

\begin{table}[h]
\centering
\caption{Composition of the 50-query TutorAccessEval dataset used in the WIP evaluation.}
\label{tab:dataset_composition}
\begin{tabular}{llr}
\toprule
Attribute & Category & Count \\
\midrule
\multirow{6}{*}{Subject} & Math & 10 \\
 & Reading & 9 \\
 & Writing & 9 \\
 & Science & 9 \\
 & Language & 9 \\
 & General & 4 \\
\midrule
\multirow{3}{*}{Grade level} & Elementary & 17 \\
 & Middle school & 21 \\
 & High school & 12 \\
\midrule
\multirow{3}{*}{Difficulty} & Easy & 20 \\
 & Medium & 13 \\
 & Hard & 17 \\
\midrule
\multirow{2}{*}{Direct-answer avoidance} & True & 25 \\
 & False & 25 \\
\bottomrule
\end{tabular}
\end{table}

The subject distribution is approximately balanced while preserving a small set of general tutoring cases. The grade-level distribution emphasizes elementary and middle-school tutoring, where scaffolding, age appropriateness, and answer-leakage avoidance are especially important. The difficulty distribution includes a substantial number of hard cases to stress-test routing decisions and escalation behavior.

\subsection{Important dataset features}
TutorAccessEval is designed around several features that are central to AI tutoring evaluation. First, each item includes an \texttt{expected\_pedagogy} field, which specifies the desired teaching strategy rather than only the desired final answer. For example, some math items ask the tutor to guide the student through inverse operations or visual reasoning; reading items may ask for contrastive examples and a check-for-understanding; writing and language items may require outlines, sentence frames, or grammar-focused scaffolding.

Second, the dataset includes explicit direct-answer-avoidance labels. This is important because good tutoring behavior is context-dependent. In some cases, such as a student asking for a short factual answer, directness is appropriate. In other cases, such as homework completion, proof-writing, translation, or essay generation, the tutor should guide the learner without completing the task for them. This distinction allows the evaluator to penalize both unwanted answer leakage and overly indirect responses when a direct answer is appropriate.

Third, each item includes risk tags that identify pedagogical or safety-relevant concerns. The most common tags include misconception, homework completion, homework dependency, writing feedback, science facts, math reasoning, reading comprehension, language grammar, and academic integrity. These tags help analyze whether failures concentrate in particular types of tutoring interactions. For instance, the case studies include academic-integrity-sensitive essay writing, direct-answer science worksheets, wholesale translation requests, and science misconception repair.

\subsection{Limitations}
TutorAccessEval is intended as a preliminary diagnostic benchmark, not a validated measure of student learning. It has several limitations. First, the dataset is small: 50 queries are sufficient for a WIP demonstration but not enough to establish robust subject-level conclusions. Second, the queries are synthetic or manually curated rather than collected from real tutoring sessions. As a result, they may not fully capture the ambiguity, emotional context, multilingual phrasing, or persistence patterns found in authentic student interactions. Third, the dataset focuses on single-turn tutoring. Many important tutoring behaviors, such as adaptive hinting, misconception tracking, and gradual fading of support, require multi-turn evaluation.

Fourth, the benchmark currently lacks demographic, accessibility, and socioeconomic context. This is a limitation for an equity-focused paper: the current access-tier AI Education Advantage Gap measures quality differences across access tiers, not across real learner populations. Fifth, the dataset does not include human teacher annotations or student learning-gain measurements. The expected pedagogy field provides a useful rubric target, but it should eventually be validated by educators. Finally, the subject distribution is broad but shallow; each domain has only a small number of examples, and some important educational settings, such as multilingual learners, special education, early literacy, and culturally grounded instruction, are underrepresented.

\subsection{Future work}
Future versions of TutorAccessEval should expand in four directions. First, the dataset should scale to hundreds or thousands of queries with balanced coverage across subjects, grade levels, difficulty levels, and tutoring risks. Second, it should include multi-turn tutoring trajectories so that systems can be evaluated on hint progression, student uptake, recovery from misunderstanding, and when to stop scaffolding. Third, future data collection should involve teachers, students, and caregivers to produce more authentic query distributions and human-validated pedagogy labels. Fourth, the benchmark should add equity-relevant slices, such as low-literacy prompts, multilingual learners, low-resource curricula, accessibility needs, and different levels of prior knowledge.

A stronger future benchmark would also include paired evaluation targets: one target for correctness and another for pedagogical appropriateness. This would allow researchers to distinguish systems that know the answer from systems that teach well. Finally, future releases should include richer metadata for cost and deployment constraints, enabling more direct evaluation of cost-aware and equity-aware routing policies in realistic educational settings.

\section{Related Work}
\label{sec:lit-review}

\subsection{Large language models for education}
LLMs are increasingly used for educational tasks such as generating explanations, giving formative feedback, supporting writing revision, and answering student questions \cite{kasneci2023chatgpt,lee2024lifecycle, wang2025llmedusurvey, chu2025llmagentseducation}. Prior work emphasizes both the promise and risk of LLMs in learning contexts: they can personalize instruction and expand access, but they can also hallucinate, over-scaffold, provide direct answers, or widen existing inequities \cite{kasneci2023chatgpt,delikoura2025risk}. These concerns motivate evaluation criteria beyond correctness, including scaffolding, age appropriateness, learner agency, and fairness.

\subsection{Cost-aware LLM routing and cascading}
LLM routing systems select among multiple models or tools to optimize performance under constraints. FrugalGPT proposes prompt adaptation, LLM approximation, and LLM cascading to reduce cost while preserving accuracy \cite{chen2023frugalgpt}. RouteLLM learns routers from preference data to select between stronger and weaker models \cite{ong2024routellm}. Other routing and cascading frameworks analyze when sequential escalation or direct model selection is most effective \cite{dekoninck2024unified}. Our work builds on this literature but changes the objective: instead of optimizing generic response quality, we optimize pedagogical quality under an access-tier quality constraint. Direct comparison to FrugalGPT and RouteLLM adapted to pedagogical tutoring---using our multi-dimensional rubric as the preference signal---is the primary planned experiment for the full version. The key architectural distinction we anticipate is that existing cascade methods optimize a single quality score and lack the multi-dimensional pedagogical rubric and the critic-rewriter stage. We cannot fairly compare against them in the current WIP without re-implementing their routing logic with pedagogical signals, which we leave to future work.

\subsection{Access-Tier Quality Gap and the AI Education Divide}
AI in education can either reduce or amplify inequalities depending on deployment context, access, and system design \cite{holmes2022ethics,kasneci2023chatgpt}. The relevant disparity is not only whether students have access to an AI tool, but whether the tool provides comparably useful learning support. In this paper, we operationalize this concern through an access-tier quality gap. Our approach is related to fairness-aware machine learning, but the unit of analysis is the tutoring interaction: for comparable student queries, how far is a budget-constrained response from the premium response in pedagogical quality? Demographic operationalization of equity---e.g., Warschauer's digital-divide framework \cite{warschauer2003technology} or Reich's analysis of MOOC equity \cite{hansen2015democratizing}---remains an important direction that we do not address in this WIP.

\section{Qualitative Evaluation Results}
\label{sec:qual-examples}

\subsection{Naive-router parity anomaly}
The naive difficulty router attains lower within-0.5-of-premium parity (72\%) than low-cost-only (78\%) despite higher mean quality (\S\ref{sec:exp}). Of the 9 queries where low-cost-only was within 0.5 of premium but the naive router was not, 5 are easy, 3 are medium, and 1 is hard---skewed toward the easy and medium queries that the naive router leaves unguarded by sending only hard queries to premium. This confirms that routing on difficulty alone raises the mean while widening the distribution, so mean quality is an insufficient parity metric.

\subsection{Failure mode analysis}
We analyze the common failure modes of each system across the 50 queries. Low-cost-only responses were most frequently flagged for weak scaffolding (18 cases), direct-answer leakage (7 cases), and incomplete responses (6 cases). Naive difficulty routing showed a similar pattern (weak scaffolding 13, direct-answer leakage 7, incomplete 6). These indicate that the low-cost model does not reliably provide high pedagogical quality. Generic cascade reduces several failure modes but still produces weak scaffolding (10 cases) and incomplete responses (5 cases). FairTutor produces the fewest logged failures: direct-answer leakage (3), incorrect facts (2), and weak scaffolding (1). Premium-only also has pedagogical failures (scoring 4.874 out of 5), including direct-answer leakage and incomplete responses, so the premium model alone does not guarantee optimal pedagogical behavior; its responses also benefit from rigorous evaluation and critic refinement.

\subsection{Case studies}
The case studies illustrate how FairTutor improves tutoring behavior. In a high-risk language-learning query, the student asked the system to write an entire five-paragraph essay comparing \emph{The Giver} and \emph{Fahrenheit 451}. low-cost-only scored 2.35 because it supplied an overly complete essay scaffold with likely fabricated or misattributed quotes and precise page numbers, creating both direct-answer leakage and factual risk. FairTutor escalated after evaluation and scored 5.00 by refusing to write the whole essay, offering thesis options, an outline, sentence frames, quote prompts, MLA guidance, and a timing plan while explicitly instructing the student to verify evidence and produce original writing.

In an elementary science query asking whether the Sun is a star, planet, or moon, low-cost-only immediately gave away the answer and included the inaccurate claim that the Sun is not orbiting anything, resulting in a score of 2.95. FairTutor scored 4.70 by giving a hint based on self-generated light, prompting the student to choose and explain, and asking follow-up questions. The remaining deduction came from an indirect clue that another star besides the Sun could reveal the answer, highlighting a useful future improvement for answer-leakage detection.

In a French translation query, the student asked for a wholesale translation of a sentence. low-cost-only scored 3.10 because it gave most of the translation directly and included a confusing explanation of agreement. FairTutor used critic rewriting and scored 5.00 by breaking the sentence into subject, verb, location, time, and idiomatic-expression chunks, explaining key grammar rules, asking check-for-understanding questions, and inviting the student to assemble the final translation. This example shows how evaluator-guided revision can transform an answer-giving response into a learning-oriented tutoring interaction.

In a science misconception query about Moon phases, low-cost-only correctly distinguished phases from eclipses but gave a misleading physical model that could accidentally demonstrate Earth-shadow eclipses rather than normal lunar phases. FairTutor scored 5.00 by directly surfacing the misconception, explaining that phases depend on how much of the Moon's sunlit half we see, and proposing a lamp-and-ball model in which the ball moves around the student's head with a slight tilt to avoid accidental eclipses. This case demonstrates that FairTutor can improve both conceptual correctness and hands-on activity design.

\end{document}